\documentclass[conference]{IEEEtran}
\IEEEoverridecommandlockouts
\usepackage{cite}
\usepackage{amsmath,amssymb,amsfonts}
\usepackage{latexsym}
\usepackage{algorithmic}
\usepackage{graphicx}
\usepackage{textcomp}
\usepackage{soul}

\usepackage{changes}
\usepackage{booktabs}
\usepackage{multirow}
\usepackage{soul}

\usepackage{xcolor}
\def\BibTeX{{\rm B\kern-.05em{\sc i\kern-.025em b}\kern-.08em
    T\kern-.1667em\lower.7ex\hbox{E}\kern-.125emX}}
\begin{document}

\makeatletter
\newcommand*\bigcdot{\mathpalette\bigcdot@{.5}}
\newcommand*\bigcdot@[2]{\mathbin{\vcenter{\hbox{\scalebox{#2}{$\m@th#1\bullet$}}}}}
\makeatother

\title{Deep Learning for Finger Vein Recognition: A Brief Survey of Recent Trend\\
\thanks{Yimin Yin and Jinghua Zhang are corresponding authors.}
}

 \author{\IEEEauthorblockN{Renye Zhang}
 \IEEEauthorblockA{\textit{School of Computer Science} \\
 \textit{Hunan First Normal University}\\
 Changsha, China  \\
 renyezhang2016@163.com}
 \and
 \IEEEauthorblockN{Yimin Yin}
 \IEEEauthorblockA{\textit{School of Mathematics and Statistics} \\
 \textit{Hunan First Normal University}\\
 Changsha, China \\
 yinyimin16@nudt.edu.cn}
 \and
 \IEEEauthorblockN{Wanxia Deng}
 \IEEEauthorblockA{\textit{College of Electronic Science} \\
 \textit{National University of Defense Technology}\\
 Changsha, China \\
 wanxiadeng@163.com}

 \and
 \IEEEauthorblockN{Chen Li}
 \IEEEauthorblockA{\textit{College of Medicine and Biological Information Engineering} \\
 \textit{Northeastern University}\\
 Shenyang, China \\
 lichen201096@hotmail.com}
 \and
 \IEEEauthorblockN{Jinghua Zhang}
 \IEEEauthorblockA{\textit{College of Intelligent Science and Technology} \\
 \textit{National University of Defense Technology}\\
 Changsha, China \\
 zhangjingh@foxmail.com}
 }

\maketitle

\begin{abstract}
Finger vein image recognition technology plays an important role in biometric recognition and has been successfully applied in many fields. Because veins are buried beneath the skin tissue, finger vein image recognition has an unparalleled advantage, which is not easily disturbed by external factors. This review summarizes 46 papers about deep learning for finger vein image recognition from 2017 to 2021. These papers are summarized according to the tasks of deep neural networks. Besides, we present the challenges and potential development directions of finger vein image recognition.
\end{abstract}


\section{Introduction}
Biometric recognition aims to identify a person based on physical features, such as fingerprint, voice, and iris~\cite{1}. With the growing requirement of digital security verification systems, the biometric recognition plays a vital role in many fields, such as online payment, security, and other fields. Compared to the traditional secure identification process, biometric recognition technology is more efficient due to its convenience and steady security. Unfortunately, several representative biometric identification technologies are struck in some bottlenecks. For instance, the fingerprint recognition rate is significantly affected by the finger surface. Besides, the fingerprints inadvertently left on things may lead to security risks. Voice recognition usually requires a relatively quiet environment. The recognition rate of iris systems is outstanding, but it requires expensive sensors. Different from above technologies, the \emph{Finger Vein Image Recognition} (FVIR) is efficient and low-cost. The finger vein is buried beneath the skin of the finger and is unique to each individual. It can be recognized through the near-infrared light~\cite{3} not the visible light which is vulnerable to external factors.

Artificial intelligence technology, especially \emph{Deep Learning} (DL) technology, has developed rapidly in recent years. Compared with transitional image processing methods, DL achieves overwhelming performance in many tasks of computer vision, such as biometric recognition~\cite{1}, biomedical image analysis~\cite{9}, and autonomous driving~\cite{10}. DL-based methods are widely used in FVIR tasks. The traditional FVIR process usually includes image capture, image data pre-processing, feature extraction, and classification or other analysis tasks. The application of DL-based methods, especially \emph{Convolutional Neural Networks} (CNNs), greatly changes the manual feature extraction process. The performance of conventional machine learning approaches is significantly influenced by feature engineering, in which the feature selection is based on human domain knowledge. Nevertheless, CNNs can extract abstract but efficient features by supervised or semi-supervised learning. The recognition process has been extremely simplified by DL-based methods.

To illustrate the recent trend and potential direction of DL-based FVIR, we conduct this brief survey. We summarize 46 related papers from 2017 to 2021, which cover different finger vein image analysis tasks, including classification, feature extraction, image enhancement, image segmentation, encryption. These papers are collected from popular academic dataset or searching engine, which mainly includes IEEE, Springer, Elsevier, and Google Scholar. We use ``finger vein image analysis'' AND (``deep learning'' OR ``neural network'' OR ``ANN” OR ``CNN''OR ``GAN'' OR ``RNN'' OR ``LSTM'') as the searching keywords. The structure of this paper is as follows: In Sec.~\ref{section2}, we briefly introduce the commonly used public datasets, some representative DL techniques, and existing survey papers. In Sec.~\ref{section4}, the related papers are summarized according to the tasks of neural networks. In Sec.~\ref{section5}, the challenges and potential directions of FVIR are talked. Finally, in Sec.~\ref{section6}, the conclusion and future work of this paper is provided.

\section{Deep learning techniques}
\label{section2}
In this section, we first introduce the commonly used public datasets. Then, the most widely used networks in FVIR are summarized. Finally, we also discuss the difference between our survey and existing papers.

\subsection{Dataset}
Dataset is critical in developing FVIR technology. The image data capacity and quality in the dataset can directly affect the performance of DL model. Through the investigation of relevant papers, the most widely used datasets are FV-USM~\cite{63}, HKPU~\cite{64}, MMCBNU-6000~\cite{65}, SDUMLA-HMT~\cite{66}, UTFVP~\cite{67}. The base information of these datasets are provided in Tab.~\ref{Tab1}.

\begin{table}[htbp!]
\caption{The base information of finger vein datasets.}
\begin{tabular}{cccc}
\hline
\textbf{Dataset} & \textbf{Data Source} & \textbf{Image Number} & \textbf{Resolution} \\ \hline
\textit{SDUMLA-HMT}       & 106                  & 3816                  & 320 $\times$ 240          \\
\textit{HKPU}             & 156                  & 3132                  & 513 $\times$ 256          \\
\textit{MMCBNU-6000}      & 100                  & 6000                  & 320 $\times$ 240          \\
\textit{FV-USM}           & 123                  & 5904                  & 640 $\times$ 480          \\
\textit{UTFVP}            & 60                   & 1440                  & 672 $\times$ 380          \\
\textit{THU-FVFDT2}       & 610                  & 2440                  & 720 $\times$ 576          \\ 
\hline
\end{tabular}
\label{Tab1}
\end{table}

\subsection{Deep Neural Networks}
AlexNet~\cite{11}, ResNet~\cite{12}, and GAN~\cite{7} are the most widely used networks in FVIR. The base characteristics of them are provided as follows:

\subsubsection{AlexNet}
\begin{figure}[htbp!]
    \centering
    \includegraphics[width = 0.45\textwidth]{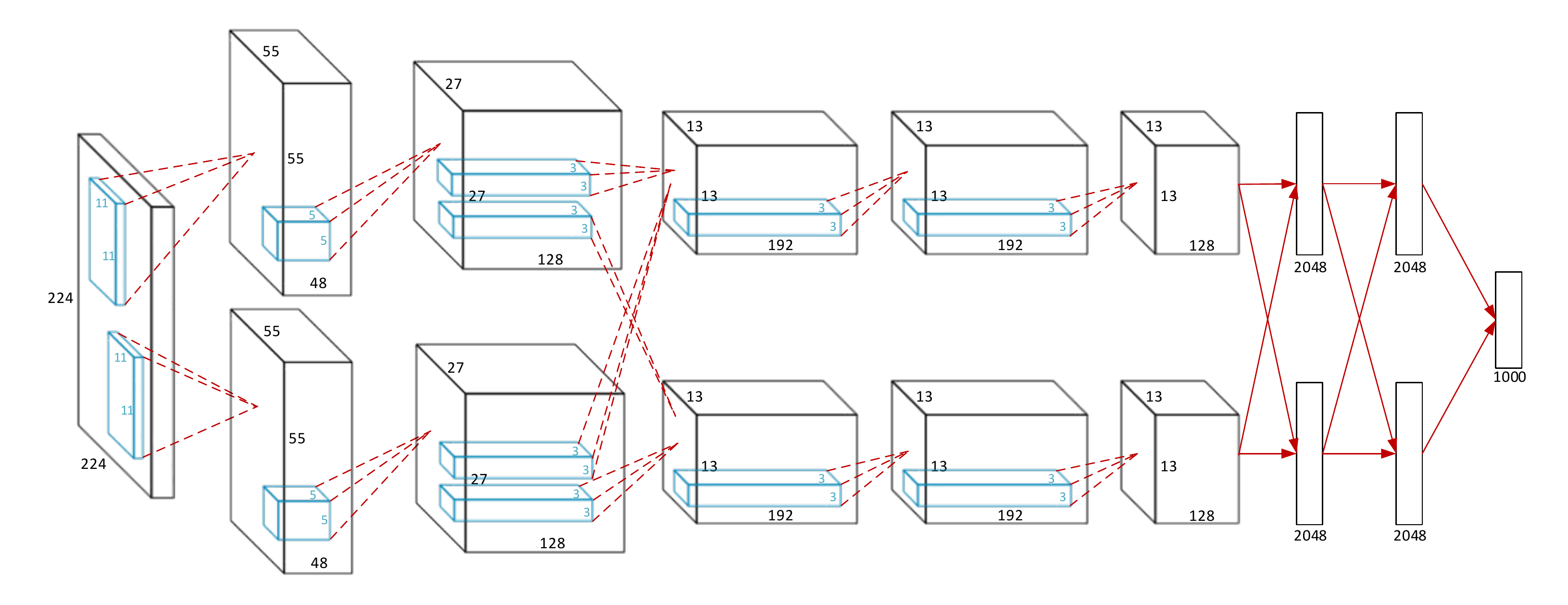}
    \caption{The structure of AlexNet}
    \label{AlexNet}
\end{figure}

AlexNet is a milestone in DL technology. Before it, the development of neural network technology was at a low ebb for many years. It is the first CNN to win the ILSVRC 2012. After AlexNet, DL starts to be the mainstream technology in many computer vision tasks~\cite{69}. The structure of AlexNet contains five convolutional layers and three max-pooling layers. Additionally, it has three fully connected layers with 4096, 4096 and 1000 neurons, respectively. Its structure is provided in Fig.~\ref{AlexNet}. This network adopts ReLU as activate function, and Dropout and data augmentation are applied to prevent over-fitting.

\subsubsection{ResNet}
\begin{figure}[!h]
    \centering
    \includegraphics[width = 0.47\textwidth]{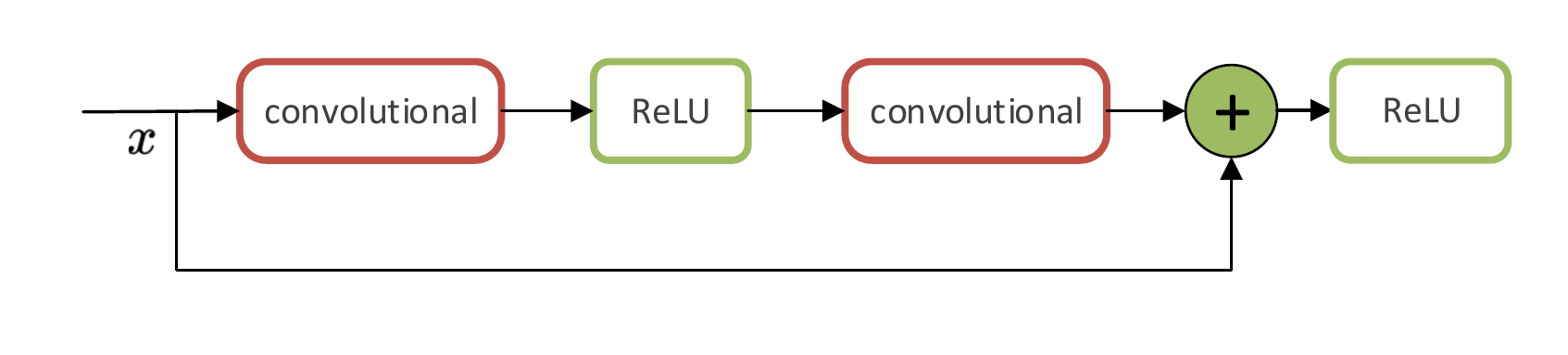}
    \caption{The residual unit structure.}
    \label{ResNet}
\end{figure}

ResNet is a deep neural network composed of several residual units connected in series. As shown in Fig.~\ref{ResNet}, the residual unit is made up of convolutional layers and a shortcut connection. The shortcut connection can effectively guarantee the back propagation of gradient. Additionally, ResNet does not contain any fully connected layer, except for the output layer. This design dramatically reduces the number of network parameters. Owing to the excellent structure of ResNet, it is widely used as the backbone in many computer vision tasks.

\subsubsection{GAN}
\begin{figure}[!h]
    \centering
    \includegraphics[width = 0.45\textwidth]{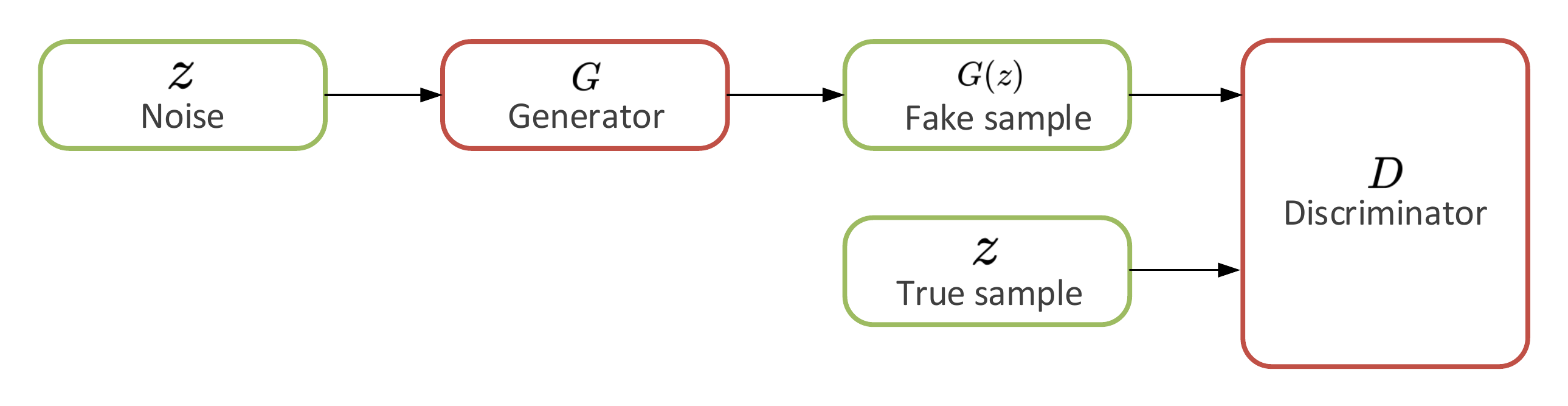}
    \caption{The basic theory of GAN}
    \label{GAN}
\end{figure}

GAN is commonly used in the finger vein image restoration task. Its structure is shown in Fig.~\ref{GAN}. Different from the neural networks used for classification, GAN is composed of the generator and discriminator. The task of generator is to generate the fake samples that can fool the discriminator. The discriminator aims to distinguish between true and false samples. This generative adversarial process can be modeled in the form of~\eqref{eq2}. $V$ is the objective function of the entire model. $D$ is discriminator, $G$ is generator, $E_{x\sim P_{data}}$ represents the true data distribution, $E_{z\sim P_{z}(z)}$ represents random noise distribution The entire formula reveals the GAN optimization process. As shown in Fig.~\ref{GAN}, the generator network receives the noisy random input, while the discriminator network receives the true sample. The output of the generator network is then fed into the discriminator, which verifies if it is a genuine sample.

\begin{equation}\label{eq2}
\begin{aligned}
\mathop{min}\limits_{Gener}\mathop{max}\limits_{Discr}V(Gener, Discr) = E_{x\sim P_{data}}[logDiscr(x)]+\\E_{z\sim P_{z}(z)}[log[1-Discr(Gener(z))]]
\end{aligned}
\end{equation}

\subsection{Analysis of Existing Reviews}

We also analyse the existing reviews about FVIR. In~\cite{13}, the main FVIR process is introduced, but the contributions of related papers are not discussed. In~\cite{5}, the amount of literature summarized is limited. \cite{15} classifies papers according to the network type, but it does not explain the tasks of networks. Different from these reviews, we conduct this survey by a detailed analysis of the neural network tasks, which can provide a new sight for the related research. Besides, our survey provides the introduction the open access datasets and the mainstream DL technologies. Additionally, a comprehensive summary table of state-of-the-art works is presented.

\section{Finger vein recognition based on deep learning}
\label{section4}
An overview of FVIR based on DL is presented in this section. According to the tasks of neural networks, the papers are divided into five  parts. In each task, the representative methods are introduced. In the end, we provide a summary table.

\subsection{Classification}
Classification is the main task in FVIR. Compared with traditional machine learning methods, DL technology shows overwhelming performance in finger vein image classification tasks. \cite{17,18,20,21,23,24,25,26,27,28,29,31,33,34,35,37,38,39,40,41,42,43,44,46,48,50,58,59} focus on the finger vein classification task based on DL. Among them, most papers adopt similar CNN-based workflow to classify the finger vein data. For example, ResNet is directly applied on the image data to perform the classification task in~\cite{23}. Similar workflows are used in~\cite{25,37,42,58}. It is worth noting that \cite{44} imports joint attention module to improve the contribution of vein patterns in feature extraction. Besides CNNs, there are some papers adopting different neural networks, such as~\cite{26} use \emph{Graph Neural Network} (GNN) to perform the classification. The intricate vein texture can be described as a graph structure. Hence, GNN can quickly distinguish different finger vein images from limited data. Additionally, \cite{24} utilizes supervised discrete hashing to increase matching speed. \cite{28} uses bias filed correction and spatial attention to optimize the CNN-based FVIR task. The module of~\cite{33} had better rotation invariance than normal CNN by using the capsule network. \cite{35} proposes a novel approach based on GAN. This method learns the joint distribution of finger vein images and pattern maps, which enhance the capacity for feature representation. \cite{46} employs the triplet loss with hard triplet online mining approach to explore the similarity between different fingers of a person.

\subsection{Feature Extraction}
The application of DL-based methods drastically alters the feature extraction process in FVIR. CNNs can automatically extract features from images. \cite{19,22,30,32,36,45,49} concentrate on DL-based feature extraction tasks. Some researches~\cite{19,30,36} adopt the analogous workflow, which usually uses CNN structure to extract features, and then adopts a traditional machine learning algorithm to analyse these features. Additionally, some studies use novel approaches to extract features, such as \cite{32} uses the\emph{Convolutional Auto-Encoder} (CAE) to learn the feature codes from finger vein images, \cite{45} proposes a capsule neural network based region of interest extraction approach for finger veins, which can represent the relationship between the part and the whole image. \cite{22} designs a lightweight two-channel network that has only three convolution layers to extract image features with an acceptable computation cost, and then support vector machine is adopted to perform the verification task. \cite{49} proposes a deep fusion of electrocardiogram and finger vein image data based multi-modal biometric authentication system. This method reaches a very high recognition accuracy.

\subsection{Image Enhancement}
Different finger vein image collection devices and user habits often lead to noisy image data in real scenarios [54], which seriously influence the performance of the DL model. To obtain high quality images sometimes, the original finger vein images must be enhanced.~\cite{51,52,53,54,55,56,57} introduce the application of DL technology in finger vein image enhancement. Among these papers, GAN has a wide range of applications. For instance, GAN is used to recover the missed vein patterns that are generated in the image capture process owing to various factors in~\cite{53}. To recover the severely damaged finger vein images,~\cite{54} proposes a modified GAN based on neighbors-based binary patterns texture loss.~\cite{55} proposes a modified DeblurGAN to increase identification performance by restoring motion blurred finger vein images to solve the problem of motion blur in FVIR. In addition to GANs, some other modules are applied in image restoration for FVIR.~\cite{51} proposes a finger vein image denoising method based on the deep CNN, the deconvolution sub-net recovers the original image based on the features, and the modified linear unit extract finger vein texture details.~\cite{52} uses a CAE to restore the venous networks of the finger vein images, thereby effectively extracting features.~\cite{56} proposes a new network architecture based on the pulse-coupled neural network to improved the finger vein image quality and increase the practicality of FVIR.
 
\subsection{Image Segmentation}
Finger vein image segmentation is an important stage in FVIR technology. The quality of segmentation has a direct impact on feature extraction and recognition. \cite{60} proposes a finger vein segmentation algorithm based on LadderNet, it can obtain abundant semantic information from vein image by concatenating the feature channels of the expanding path and contracting path in the network. Additionally, the parameters of normal finger vein segmentation networks are overabundant, which makes they are challenging to use in mobile terminals. To overcome this problem, \cite{61} proposes a lightweight real-time segmentation network in FVIR based on the embedded terminal. The performance of this network is not inferior to more complex networks and satisfied the needs of embedded mobile terminals.

\subsection{Encryption}
Since biometric information is irreplaceable and unique for everyone, once the original biometric information is stolen, it may cause irreversible loss. To protect the privacy of users more effectively, encryption methods are used in FVIR. This technology masks the biometric information in the image by encrypting the original image. Even if the finger vein image is stolen, criminals cannot obtain valid information from it. However, the biggest challenge of encryption is how to keep the performance of the recognition system while protecting biometric data. To address this issue, \cite{62} presented a novel FVIR algorithm by using a secure biometric template scheme based on DL and random projections. This algorithm randomly generates a secured template for the original biometric message by random projections. \cite{47} proposes a deep CAE structure to reduce the dimension of the feature space and introduced the Biohashing algorithm to generate protected templates based on the features that were extracted at the CAE.

\begin{table*}[htbp!]
\renewcommand{\arraystretch}{1.2}
\setlength\tabcolsep{2.5pt}
\caption{The summary table of surveyed papers. For short, Task, Dataset, Reference, Network, Performance, Classification, Feature extraction, Image enhancement, Image segmentation, encryption, SDUMLA-HMT, HKPU, MMCBNU-6000, FV-USM, UTFVP, THU-FVFDT2, SCUT, IDIAP, PLUSVein-FV3, private, VeinECG, ISPR, NJUST-FV, Accuracy, Equal Error Rate, Peak Signal-to-Noise Ratio, Structural Similarity, Presentation Classification Error Rate and Bona-fide Presentation Classification Error Rate are abbreviated to T, D, Ref, Net, Perf, C*, F*, IE*, S*, E*, SD*, HK*, MM*, US*, UT*, TH*, SC*, ID*, PL*, PR*, VE*, IS*, NJ*, ACC, EERE, PSNR, SSIM, APCER and BPCER.} 
\begin{tabular}{|c|c|c|c|c|c|c|c|c|c|c|c|c|c|c|}
\hline
\textbf{T.}                            & \textbf{D.}                    & \textbf{Ref.} & \textbf{Net.}   & \textbf{Perf.(\%)}                                            & \textbf{T.}                            & \textbf{D.}                   & \textbf{Ref.} & \textbf{Net.}       & \textbf{Perf.(\%)}                                             & \textbf{T.}                             & \textbf{D.}                   & \textbf{Ref.} & \textbf{Net.}       & \textbf{Perf.(\%)}                                              \\ \hline
\multirow{27}{*}{\textit{\textbf{C*}}} & \multirow{15}{*}{\textit{SD*}} & \cite{23}     & ResNet-101      & EERE 3.3653                                                    & \multirow{25}{*}{\textit{\textbf{C*}}} & \multirow{3}{*}{\textit{MM*}} & \cite{44}      & JAFVNet             & EERE 0.23                                                       & \multirow{10}{*}{\textit{\textbf{F*}}}  & \textit{SD*}                  & \cite{45}     & Capsule Network     & ACC 97.5                                                        \\ \cline{3-5} \cline{8-10} \cline{12-15} 
                                       &                                & \cite{33}      & Capsule Network & ACC 100                                                       &                                        &                               & \cite{43}      & ResNet              & \begin{tabular}[c]{@{}c@{}}EERE 0.090\\ CIR 99.667\end{tabular} &                                         & \textit{HK*}                  & \cite{19}      & Improved CNN & ACC 87.08                                                       \\ \cline{3-5} \cline{8-10} \cline{12-15} 
                                       &                                & \cite{24}      & Light CNN       & EERE 0.1497                                                    &                                        &                               & \cite{50}      & two-branch CNN      & EERE 0.17                                                       &                                         & \textit{MM*}                  & \cite{22}      & two-stream network  & EERE 0.10                                                        \\ \cline{3-5} \cline{7-10} \cline{12-15} 
                                       &                                & \cite{31}      & RefineNet       & EERE 2.45                                                      &                                        & \multirow{7}{*}{\textit{US*}} & \cite{20}      & Improved CNN        & EERE 1.42                                                       &                                         & \multirow{3}{*}{\textit{US*}} & \cite{32}      & CAE                 & \begin{tabular}[c]{@{}c@{}}ACC 99.49\\ EERE 0.16\end{tabular}    \\ \cline{3-5} \cline{8-10} \cline{13-15} 
                                       &                                & \cite{35}      & GAN             & EERE 0.94                                                      &                                        &                               & \cite{25}      & DBN                 & ACC 97.4                                                       &                                         &                               & \cite{36}      & PCANet              & ACC 99.49                                                       \\ \cline{3-5} \cline{8-10} \cline{13-15} 
                                       &                                & \cite{26}      & GNN             & ACC 99.98                                                     &                                        &                               & \cite{40}      & Sub-CNN             & ACC 95.1                                                       &                                         &                               & \cite{45}      & Capsule Network     & ACC 99.7                                                        \\ \cline{3-5} \cline{8-10} \cline{12-15} 
                                       &                                & \cite{37}      & DenseNet-201    & EERE 0.54                                                      &                                        &                               & \cite{42}      & Improved CNN        & \begin{tabular}[c]{@{}c@{}}ACC 97.95\\ EERE 1.070\end{tabular}  &                                         & \textit{TH*}                  & \cite{36}      & PCANet              & ACC 100                                                         \\ \cline{3-5} \cline{8-10} \cline{12-15} 
                                       &                                & \cite{38}      & DenseNet-161    & EERE 2.35                                                      &                                        &                               & \cite{43}      & ResNet              & EERE 0.091                                                      &                                         & \textit{ID*}                  & \cite{30}      & VGG-16              & EERE 0.0000                                                      \\ \cline{3-5} \cline{8-10} \cline{12-15} 
                                       &                                & \cite{39}      & DenseNet-161    & EERE 1.65                                                      &                                        &                               & \cite{44}      & JAFVNet             & EERE 0.49                                                       &                                         & \textit{VE*}                  & \cite{49}      & Improved CNN        & EERE 0.12                                                        \\ \cline{3-5} \cline{8-10} \cline{12-15} 
                                       &                                & \cite{28}      & ResNet-50       & ACC 99.53                                                     &                                        &                               & \cite{29}      & NASNet              & ACC 98.89                                                      &                                         & \textit{IS*}                  & \cite{30}      & VGG-16              & EERE 0.0311                                                      \\ \cline{3-5} \cline{7-15} 
                                       &                                & \cite{43}      & ResNet          & EERE 2.137                                                     &                                        & \multirow{3}{*}{\textit{UT*}} & \cite{33}      & Capsule Network     & ACC 94                                                         & \multirow{11}{*}{\textit{\textbf{IE*}}} & \multirow{3}{*}{\textit{SD*}} & \cite{55}      & GAN                 & EERE 5.270                                                       \\ \cline{3-5} \cline{8-10} \cline{13-15} 
                                       &                                & \cite{44}      & JAFVNet         & EERE 1.18                                                      &                                        &                               & \cite{46}      & SqueezeNet          & EERE 2.5                                                        &                                         &                               & \cite{56}      & PCNN                   & -                                                               \\ \cline{3-5} \cline{8-10} \cline{13-15} 
                                       &                                & \cite{46}      & SqueezeNet      & EERE 2.7                                                       &                                        &                               & \cite{31}      & U-Net               & EERE 0.64                                                       &                                         &                               & \cite{57}      & GAN                 & EERE 0.87                                                        \\ \cline{3-5} \cline{7-10} \cline{12-15} 
                                       &                                & \cite{58}      & GoogleNet       & ACC 92.22                                                     &                                        & \multirow{2}{*}{\textit{TH*}} & \cite{35}      & GAN                 & \begin{tabular}[c]{@{}c@{}}ACC 98.52\\ EERE 1.12\end{tabular}   &                                         & \multirow{3}{*}{\textit{HK*}} & \cite{51}      & Improved CNN        & PSNR 29.638                                                     \\ \cline{3-5} \cline{8-10} \cline{13-15} 
                                       &                                & \cite{50}      & two-branch CNN  & EERE 0.94                                                      &                                        &                               & \cite{28}      & ResNet-50           & ACC 98.64                                                      &                                         &                               & \cite{55}      & GAN                 & EERE 4.536                                                       \\ \cline{2-5} \cline{7-10} \cline{13-15} 
                                       & \multirow{9}{*}{\textit{HK*}}  & \cite{20}      & Improved CNN    & EERE 2.70                                                      &                                        & \multirow{3}{*}{\textit{SC*}} & \cite{17}     & FPNet               & EERE 0.00                                                       &                                         &                               & \cite{56}      & PCNN                & -                                                               \\ \cline{3-5} \cline{8-10} \cline{12-15} 
                                       &                                & {\cite{23}}      & ResNet-101      & EERE 1.0799                                                    &                                        &                               & \cite{41}      & FVRASNet            & EERE 2.02                                                       &                                         & \textit{MM*}                  & \cite{53}      & GAN                 & EERE 5.66                                                        \\ \cline{3-5} \cline{8-10} \cline{12-15} 
                                       &                                & \cite{33}      & Capsule Network & ACC 88                                                        &                                        &                               & \cite{44}      & JAFVNet             & EERE 0.86                                                       &                                         & \textit{US*}                  & \cite{53}      & GAN                 & EERE 2.37                                                        \\ \cline{3-5} \cline{7-10} \cline{12-15} 
                                       &                                & \cite{34}      & LSTM            & EERE 0.95                                                      &                                        & \multirow{2}{*}{\textit{ID*}} & \cite{17}      & FPNet               & EERE 0.25                                                       &                                         & \multirow{2}{*}{\textit{PR*}} & \cite{52}      & CAE                 & EERE 0.16                                                        \\ \cline{3-5} \cline{8-10} \cline{13-15} 
                                       &                                & \cite{38}      & DenseNet-161    & EERE 0.33                                                      &                                        &                               & \cite{41}      & FVRASNet            & EERE 4.26                                                       &                                         &                               & \cite{54}      & GAN                 & \begin{tabular}[c]{@{}c@{}}PSNR 30.42\\ SSIM 98.85\end{tabular} \\ \cline{3-5} \cline{7-10} \cline{12-15} 
                                       &                                & \cite{39}      & DenseNet-161    & EERE 0.05                                                      &                                        & \multirow{2}{*}{\textit{PL*}} & \cite{46}      & SqueezeNet          & EERE 2.4                                                        &                                         & \textit{NJ*}                  & \cite{56}      & PCNN                & -                                                               \\ \cline{3-5} \cline{8-15} 
                                       &                                & \cite{43}      & ResNet          & EERE 0.277                                                     &                                        &                               & \cite{48}      & Triplrt-SqNet       & EERE 3                                                          & \multirow{4}{*}{\textit{\textbf{S*}}}   & \multirow{2}{*}{\textit{SD*}} & \cite{60}      & LadderNet           & ACC 92.44                                                       \\ \cline{3-5} \cline{7-10} \cline{13-15} 
                                       &                                & \cite{46}      & SqueezeNet      & EERE 3.7                                                       &                                        & \multirow{3}{*}{\textit{PR*}} & \cite{21}      & AlexNet             & \begin{tabular}[c]{@{}c@{}}APCER 0\\ BPCER 0\end{tabular}      &                                         &                               & \cite{61}      & DintyNet            & ACC 91.93                                                       \\ \cline{3-5} \cline{8-10} \cline{12-15} 
                                       &                                & \cite{59}      & Improved CNN    & ACC 91.19                                                     &                                        &                               & \cite{25}      & DBN                 & ACC 97.8                                                       &                                         & \multirow{2}{*}{\textit{MM*}} & \cite{60}      & LadderNet           & ACC 93.93                                                       \\ \cline{2-5} \cline{8-10} \cline{13-15} 
                                       & \multirow{3}{*}{\textit{MM*}}  & \cite{33}      & Capsule Network & ACC 100                                                       &                                        &                               & \cite{27}      & LSTM                & ACC 99.13                                                      &                                         &                               & \cite{61}      & DintyNet            & ACC 90.90                                                       \\ \cline{3-15} 
                                       &                                & \cite{26}      & GNN             & ACC 99.98                                                     & \multirow{2}{*}{\textit{\textbf{F*}}}  & \multirow{2}{*}{\textit{SD*}} & \cite{22}      & two-channel network & EERE 0.47                                                       & \multirow{2}{*}{\textit{\textbf{E*}}}   & \textit{UT*}                  & \cite{47}      & CAE                 & EERE 0.7                                                         \\ \cline{3-5} \cline{8-10} \cline{12-15} 
                                       &                                & \cite{42}      & Improved CNN    & \begin{tabular}[c]{@{}c@{}}ACC 99.05\\ EERE 0.503\end{tabular} &                                        &                               & \cite{36}      & PCANet              & ACC 98.19                                                      &                                         & \textit{PR*}                  & \cite{62}      & CAE                 & \begin{tabular}[c]{@{}c@{}}GAR 96.9\\ FAR 1.5\end{tabular}      \\ \hline
\end{tabular}
\end{table*}

\section{Challenge and potential direction}
\label{section5}
Compared with traditional biometric technology, FVIR has unparalleled advantages but it still faces some challenges, especially during image capture~\cite{54},~\cite{68},~\cite{41}, which contains uneven illumination, light scattering in finger tissue, inappropriate ambient temperature, image displacement, presentation attacks, shading, etc. All of the aforementioned challenges have varying degrees of impact on the performance of FVIR. To overcome these challenges, \cite{17,30,21,56,57,53,41,54} try to propose solutions from different aspects. However, these technical difficulties have not been completely solved, and they will remain the focus of FVIR in the future.

Besides, FVIR also has some potential directions. For instance, FVIR generally needs to be implemented on lightweight portable mobile terminals. However, most of the deep neural networks are not suitable for this kind of device. Therefore, DL-based FVIR faces some difficulties in practice. Knowledge distillation~\cite{73} can be utilized to overcome this challenge. Knowledge distillation can greatly condense complicated networks by teaching a small student model from a large model. This technology can greatly improve the application ability of FVIR in reality. 

Furthermore, one of the conveniences of FVIR is that even if one finger is in an accident, the other fingers can still be used for identification. But registering ten fingers simultaneously in an identification system is a hassle for users. Therefore, it is necessary to explore whether the finger veins of the ten fingers of the same individual are similar in future FVIR research work. If there is some connections between different finger veins of the same person and they can be identified by FVIR systems, it will take the convenience of FVIR systems to a new level. Although \cite{37, 46} focus on this problem, they still have some limitations.\cite{37} considers the connection between the veins of different fingers of the same person is too weak to perform the recognition. \cite{46} utilizes the triplet loss with hard triplet online mining for FVIR. This strategy successfully verified that symmetric fingers (the same sort of finger but from opposite hands in the same individual) have enough similarities to be recognized. The similarities of other asymmetric fingers is also proved in ~\cite{46}, but the proposed recognition system is still unable to effectively identify these asymmetric finger veins. Therefore, related work can still be further explored in the future.

\section{Conclusion and future work}
\label{section6}
In this brief survey, we summarize the DL technology for FVIR. First, we introduce the base information of widely used public datasets and some popular CNN structures. After that, we summarize 46 related research literature of FVIR based on DL from 2017 to 2021, and classify them according to the tasks of neural networks, which includes classification, feature extraction, image enhancement, image segmentation, encryption. Finally, we discuss the current challenges and development directions of FVIR. From this review, it can be found that the tasks of neural networks are diverse in FVIR, and compared to other biometric recognition systems, FVIR has unique advantages. In the future, we will investigate more literature and DL techniques in FVIR to propose a comprehensive review.



\end{document}